\pdfoutput=1

\documentclass[12pt]{article}

\usepackage{graphicx}
\usepackage{url}

\usepackage{authblk}

\usepackage{amssymb}
\usepackage{amsmath}

\usepackage[table]{xcolor}
\newcommand{\change}[1]{#1}

\usepackage[inline=true, margin=false]{fixme}
\usepackage{hyperref}

\title{Predicting Kidney Transplant Survival using Multiple Feature Representations for HLAs}

\author[1]{Mohammadreza Nemati}
\author[1]{Haonan Zhang}
\author[1]{Michael Sloma}
\author[2]{Dulat Bekbolsynov}
\author[3]{Hong Wang}
\author[2]{Stanislaw Stepkowski}
\author[1]{Kevin S. Xu\thanks{Corresponding author:  \url{kevin.xu@utoledo.edu}}}

\affil[1]{Department of Electrical Engineering and Computer Science, University of Toledo, Toledo, OH, USA}
\affil[2]{Department of Medical Microbiology and Immunology, University of Toledo}
\affil[3]{Department of Engineering Technology, University of Toledo}

\begin{document}

\maketitle

\begin{abstract}
Kidney transplantation can significantly enhance living standards for people suffering from end-stage renal disease.
A significant factor that affects graft survival time (the time until the transplant fails and the patient requires another transplant) for kidney transplantation is the compatibility of the Human Leukocyte Antigens (HLAs) between the donor and recipient. 
In this paper, we propose 4 new biologically-relevant feature representations for incorporating HLA information into machine learning-based survival analysis algorithms.
We evaluate our proposed HLA feature representations on a database of over 100,000 transplants and find that they improve prediction accuracy by about 1\%, modest at the patient level but potentially significant at a societal level. 
Accurate prediction of survival times can improve transplant survival outcomes, enabling better allocation of donors to recipients and reducing the number of re-transplants due to graft failure with poorly matched donors.

\end{abstract}

\section{Introduction}
Kidney transplantation is the therapy of choice for many people suffering from end-stage renal disease (ESRD). A successful kidney transplant can enhance a patient's living standards and diminish the patient's risk of dying. Although allograft (organ or tissue transplanted from one individual to another) and patient survival have improved because of new surgical technologies and effective immunosuppression,  a transplant is not a lifetime treatment. Allografts, or simply grafts, will stop functioning over time \cite{sellares2012understanding}, 
requiring re-transplantation for the patient after graft failure. There is a significant societal demand for kidney transplants, with over 90,000 people on the waiting list in the United States \change{alone.}

The time to graft failure or \emph{graft survival time} is determined by a variety of factors, including the age, race, and overall health of the donor and recipient. 
The \emph{compatibility} of the donor and recipient also plays a key role, particularly with respect to their Human  Leukocyte Antigens (HLAs) \cite{opelz1999hla}.
Prior research has demonstrated that the number of mismatches (MM) between donor and recipient HLAs can significantly affect the graft survival time \cite{casey2015rethinking, foster2014impact, kwon2004impact}.

In this paper, we aim to \emph{predict} the graft survival time for a transplant given a variety of covariates on the donor and recipient. 
We propose \emph{multiple feature representations} for incorporating HLA information into survival analysis models. 
By building a base model without HLA information and then comparing to models that contain more detailed representations of HLAs, we can identify whether the HLA information can improve prediction accuracy.  

Our main contribution is 4 new feature representations for HLA types and pairs that account for \emph{biological mechanisms} behind HLA compatibility, differences in categorization of HLAs, and differences in the way categorical variables are treated in different survival analysis models. 
We find that incorporating HLA information can improve the accuracy of predicted graft survival time by about 1\%. 
While this is a modest improvement for an individual patient, it could translate to significant improvements at the societal level by increasing graft survival times, thus enabling more transplant \emph{recipients} with the same number of donors, and potentially reducing the size of the waiting list.

We first reported preliminary results from this paper in the conference publication \cite{nemati21}. 
This paper extends those results with the following new contributions:
\begin{itemize}
    \item We propose 2 new target encoding approaches for HLA types designed to improve prediction accuracy with random survival forests and other tree-based survival analysis models.
    
    \item We repeat our experiments over 10 different random \change{train, validation, and test splits, unlike our preliminary results in \cite{nemati21} that used only a single split. In doing so, we establish statistical significance for our main findings and can further conclude that the observed improvements from incorporating HLA features, while small, are unlikely to be due to chance alone.}
    
    \item We present a detailed comparison of prediction accuracy when including also post-transplant covariates for prediction.
    
    \item \change{We add a more detailed discussion of the potential clinical significance and impact of our biologically-relevant HLA feature representations.}
\end{itemize}

\section{Background and Motivation}
Chronic kidney disease (CKD) is a public health issue and a general term for heterogeneous disorders affecting a kidney's function, which may lead to ESRD. 
According to 2019 reports of United States Renal Data System, CKD affects at least 10\% of adults in the U.S., 
with nearly 750,000 Americans requiring kidney transplantation. In the absence of kidney donors, life support therapy for these patients is associated with exorbitant morbidity, mortality, and tremendous financial burden. 
Successful kidney transplantation may save about \$55,000 per year in Medicare costs for every functioning transplant~\cite{salomon2015ast}. 

Unfortunately, the waiting list for kidney transplantation continues to grow. 
Based on 2019 OPTN data, over 41,000 new patients were added to the kidney transplant waiting list, while only 23,401 total transplants were performed, with 11\% of them being patients returning to the waiting list due to previous transplant failure. 
These numbers highlight the need to improve transplant survival in kidney transplant recipients.

\subsection{Human Leukocyte Antigens (HLAs)}
\label{sec:hla}
HLAs are a category of surface proteins encoded in a distinct gene cluster \cite{campbell1993map}. 
These HLAs, which are highly polymorphic, play a fundamental role in the body's immune system function. 
In organ transplantation, donor HLAs are also recognized as foreign to be attacked by the recipient’s immune system~\cite{horton2004gene}. 
Each human inherits 2 copies (1 maternal and 1 paternal) of each HLA gene. In the cluster, 3 specific loci, HLA-A, -B and -DR, are of utmost clinical significance for kidney transplantation. Thus, 6 HLAs (2 copies of each of HLA-A, -B, and -DR) are routinely typed in the clinic. 
An HLA is typically represented by the locus and a 2-digit number such as A1. 
This representation is known as the HLA serological type; we refer to it as just the HLA type in this paper.
For example, a donor may have the following 6 HLA types: A3, A9, B5, B7, DR5, and DR6. 

It is important to note that the 2-digit numbers are just categories and not actual numerical values. 
For example, A9 and A10 are not necessarily more similar than A9 and A3 despite the numbers 9 and 10 being closer than 9 and 3. 
Furthermore, as HLA typing methods evolved, some crudely defined antigens (broads) were found to be groups of finer, previously unseen antigens (splits).
For example, the splits and associated antigens of the broad antigen HLA-A9 are A23 and A24. 
Thus, some instances of the splits A23 and A24 may be coded as the broad A9 in transplant databases. 

The clinical importance of HLA stems from the sheer polymorphism~\cite{robinson2020ipd}, resulting in donor HLAs being in most instances different from recipient HLAs.
Each HLA type present in the donor but not in the recipient leads to an HLA mismatch (MM). 
There may be 0 to 6 HLA-A/B/DR MM between a donor and recipient, with higher MM generally resulting in shorter graft survival times \cite{opelz1999hla}.
In addition to the number of mismatches, prior work has also demonstrated that specific HLAs of the donor and recipient can also impact the outcome of kidney allograft survival~\cite{cheigh1977renal}.

\subsection{Survival Analysis}

\begin{figure}[t]
    \centering
    \includegraphics[width = 4in]{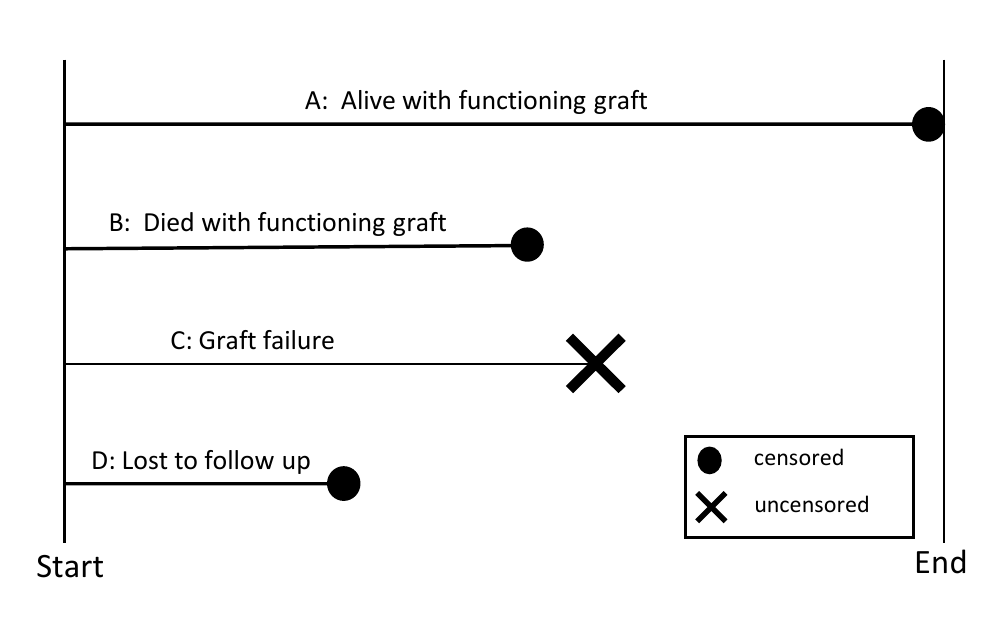}
    \caption{Examples of right censoring in the kidney transplantation setting we consider.}
    \label{fig:censoring}
\end{figure}

Survival analysis is a well-established technique in statistics used to predict time to an event of interest during a specific observed time interval. It is a form of regression where the objective is to predict the survival time, i.e.~the time until an event of interest occurs.   For many data points, however, the exact time of the event is unknown due to \emph{censoring}, and thus,
standard regression models are not well-suited to handle such time-to-event problems. 
Right censoring is the most common form and applies to our survival time prediction problem. Although a subject's status is known at the beginning of the study, the subject's event might not be observed. 
Some of the censoring conditions in our kidney transplantation problem setting are shown in Figure~\ref{fig:censoring}.

Many survival analysis algorithms have been proposed to handle censored data---we refer readers to the survey \cite{wang2019machine}.
We consider 3 machine learning-based survival analysis algorithms in this paper, which we describe in Section \ref{sec:algorithms}.

\section{Data Description}
This study uses data from the Scientific Registry of Transplant Recipients (SRTR) and includes data on all donors, wait-listed candidates, and transplant recipients in the U.S., submitted by the members of the Organ Procurement and
Transplantation Network (OPTN). 

\paragraph{Inclusion Criteria}
We acquired $469,711$ anonymous cases on all kidney transplants between 1987 and 2016 from the registry. 
We apply the following inclusion criteria to the data. 
We consider only transplants with deceased donors, recipients aged 18 years or older, and only candidates who are receiving their first transplant. 
We include only transplants between 2000 and 2016 due to the introduction of new therapy regimes and a new kidney allocation system~\cite{ashby2011transplanting} around the year 2000. 
Finally, we include only recipients with peak Panel Reactive Antibody (PRA) less than 80 percent
since patients with high PRA levels experience increased acute rejection rate and graft failure~\cite{schwaiger2016deceased}.
After applying the inclusion criteria, subjects with missing values in any basic covariates except cold ischemia time (see Section \ref{sec:basic}) or HLAs are removed from the study. 
After the preceding stages, $106,372$ transplants remain for the purpose of developing predictive models, with 74.6 percent of them being censored. 

\paragraph{Target Variable}
There are three primary endpoints (targets) in survival analyses for kidney transplantation: patient survival, all-cause graft loss, and death-censored graft loss. In the SRTR database, data on patient survival was compiled based on reports from transplant centers, as well as the Centers for Medicare and Medicaid Services and Social Security Administration's Death Master File~\cite{leppke2013scientific}. Record of patient death and patient death date from any of these sources was used to define the patient survival variable. 

We use \emph{death-censored graft loss} as the clinical endpoint (prediction target) in this study. 
This means that patients who died with a functioning graft are treated as censored since they did not exhibit the event of interest (graft loss), as shown in example B in Figure \ref{fig:censoring}. 
Graft loss is determined based on the record of either graft failure, return to maintenance dialysis, re-transplant, or listing for re-transplant.
For censored instances, the censoring date is defined to be the last follow-up date.

\section{Methods and Technical Solutions}
\begin{figure}[t]
    \centering
    \includegraphics[width = \linewidth]{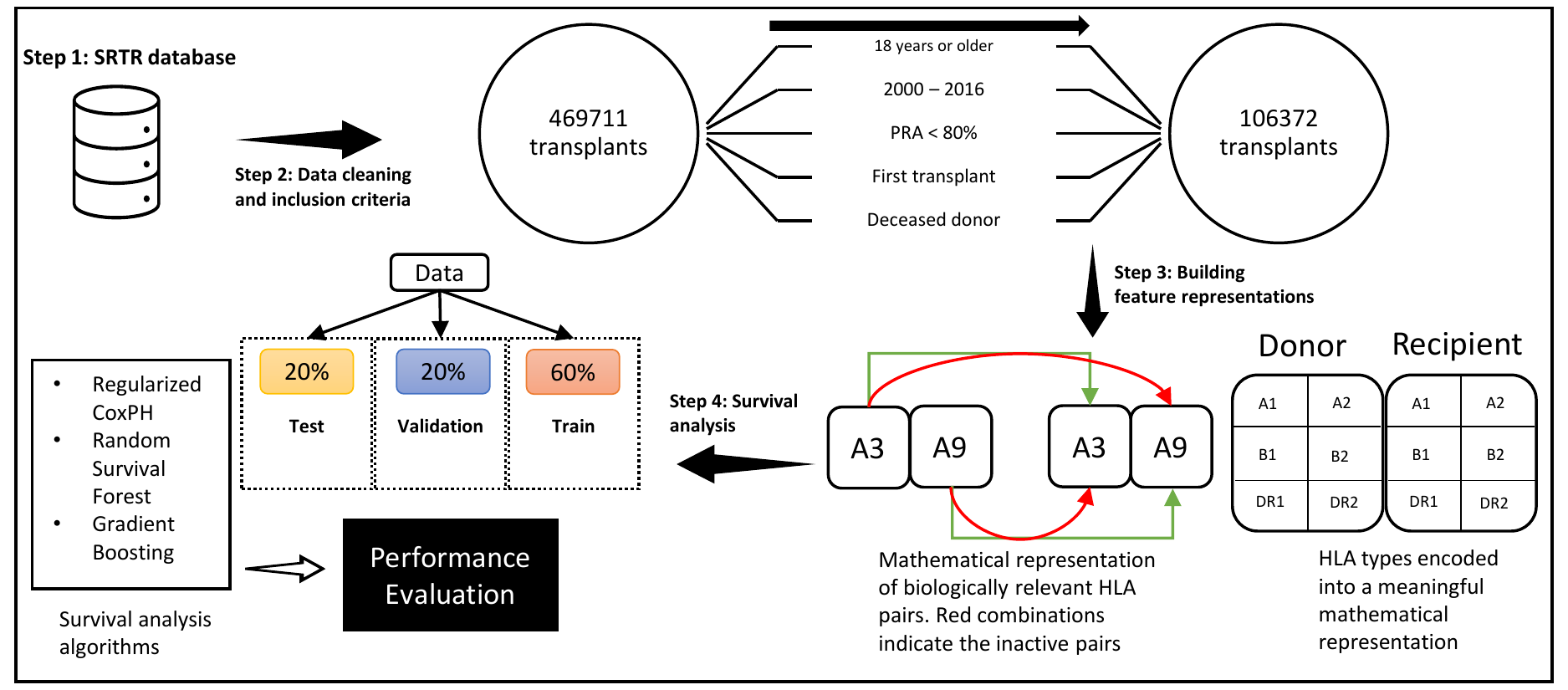}
    \caption{Illustration of data processing and survival analysis pipeline.}
    \label{fig:pipeline}
\end{figure}

\paragraph{Research Questions} 
We pose two main research questions in this study. 
First, does incorporating donor and recipient HLA information into a graft survival time predictor improve prediction accuracy? 
If so, what type of representation for the HLA information results in the highest prediction accuracy?
We first describe the different HLA feature representations we propose in Section \ref{sec:features} and then discuss the survival analysis algorithms we use in Section \ref{sec:algorithms}.
Our data processing pipeline is shown in Figure \ref{fig:pipeline}.

\subsection{Feature Representations}
\label{sec:features}
\begin{table}[t]
\centering
\caption{Number of covariates in each feature set in pre- and post-transplant settings.}
\label{tab:covariate_count}
\begin{tabular}{ccc}
\hline
                                & \multicolumn{2}{c}{Number of covariates} \\
Feature Set                     & Pre-transplant     & Post-transplant \\
\hline
Basic                           &23                  &29          \\
\change{Basic + MM (total)}     &24                  &30          \\
\change{Basic + MM (A-B-DR)}    &26                  &32          \\ 
\change{Basic + Types (binary)} &252                 &258          \\
\change{Basic + Types (target)} &29                  &35                \\
\change{Basic + Pairs}          &3,661               &3,667          \\
\change{Basic + Frequent pairs} &227                 &236          \\ 
All                             &3,894               &3,900          \\
\hline
\end{tabular}
\end{table}

We consider 8 different feature sets ranging from 23 to 3,900 covariates. Some of the covariates are \emph{pre-transplant covariates}, meaning that they are available prior to the transplant time, while others are post-transplant covariates, available only at the time of transplant or after a transplant has been performed and the patient has been discharged. 
The number of covariates for each feature set is shown in Table \ref{tab:covariate_count}.

We first consider prediction using only the pre-transplant covariates, as they can be used to predict graft survival prior to the transplant being performed and could potentially be used in the process of matching donors and recipients.
We also consider prediction using both pre- and post-transplant covariates, which should be more accurate and can still be useful to a clinician. 

\subsubsection{Basic Features}
\label{sec:basic}
Pre-transplant basic features consist of age, sex, race, and body mass index (BMI). 
Race is encoded using a one-hot representation. 
The post-transplant covariates employed are donor and recipient serum creatinine levels at the time of transplant, recipient serum creatinine at discharge time, whether the patient needs dialysis within the first week of the transplant, and the cold ischemia time (CIT), which denotes the amount of time the kidney was preserved after the blood supply has been cut off.
Missing values for CIT were imputed using the mean over all other transplants. 
There are a total of 23 and 29 features for the pre- and post-transplant settings, respectively. 

\subsubsection{HLA Mismatches}
We first consider the number of mismatches (MM) between donor and recipient, which has been found to be a significant factor in the time to graft failure. 
We consider two possible representations: the total number of MM (0 to 6), as well as the separate A-B-DR MM (0 to 2 each). 
These result in 1 and 3 features appended to the basic features, respectively.

\subsubsection{HLA Types}
\label{sec:hla_types}
We consider directly encoding the HLA types of the donor and recipient.
The digits in an HLA type should be treated as categories and not numeric values, e.g.~A2 and A1 differing by 1 does not imply that they are more similar than A2 and A23.

\paragraph{Binary Encoding}
One focus of our study is to address the methodological challenges arising from HLA broad and split antigens. 
We propose to encode HLA types using a binary one-hot-like encoding that also maps splits back to broads so that a split like A23 has a one in both the columns for A23 and A9. 
We encode donors and recipients separately so that each transplant has at least 12 ones (6 donor, 6 recipient), and possibly more due to splits. 
\change{An example of the binary encoding applied to donor and recipient HLA types for the HLA-A locus is shown in Figure \ref{fig: HLA type and pair encoding}.}
This encoding results in 229 features appended to the basic features.

\paragraph{Target Encoding}
One disadvantage to the proposed binary encoding is that it converts 12 categorical features for the donor and recipient HLA types into 229 binary features. 
This may have an adverse effect on the accuracy of tree-based \change{models. (Indeed, we observe that it leads to a decrease in the prediction accuracy of  the random survival forest but not that of the Coxnet or gradient boosting, as we show in Section \ref{sec:results_features}.)
To mitigate this disadvantage, we propose an alternative encoding: a} \emph{target encoding} approach that encodes HLA types with a much lower dimensionality, resulting in 6 real-valued features as opposed to 229 binary features. 
\change{This target encoding approach was not considered in our preliminary results in \cite{nemati21}.}

The proposed technique relies on a transformation that maps each category of a high-cardinality categorical variable to the target variable's probability estimate. In a typical supervised learning setting, the numerical representation corresponds to the target's expected value given the categorical feature's category \cite{micci2001preprocessing}. 
In our setting, however, we have two additional challenges. 
The first is censoring, which prevents us from observing the time to event for the majority of instances. 
Secondly, each person inherits two copies of each HLA, one paternal and one maternal. 
These two HLA types are typically stored as two different covariates, e.g.~DON\_A1 and DON\_A2 denoting the two HLA-A types that the donor possesses. 
However, the ordering of the two types does not matter, so that (DON\_A1, DON\_A2) $= (3, 9)$ and (DON\_A1, DON\_A2) $= (9, 3)$ both denote a donor possessing A3 and A9.  

\begin{figure}[t]
    \centering
    \includegraphics[width = \linewidth]{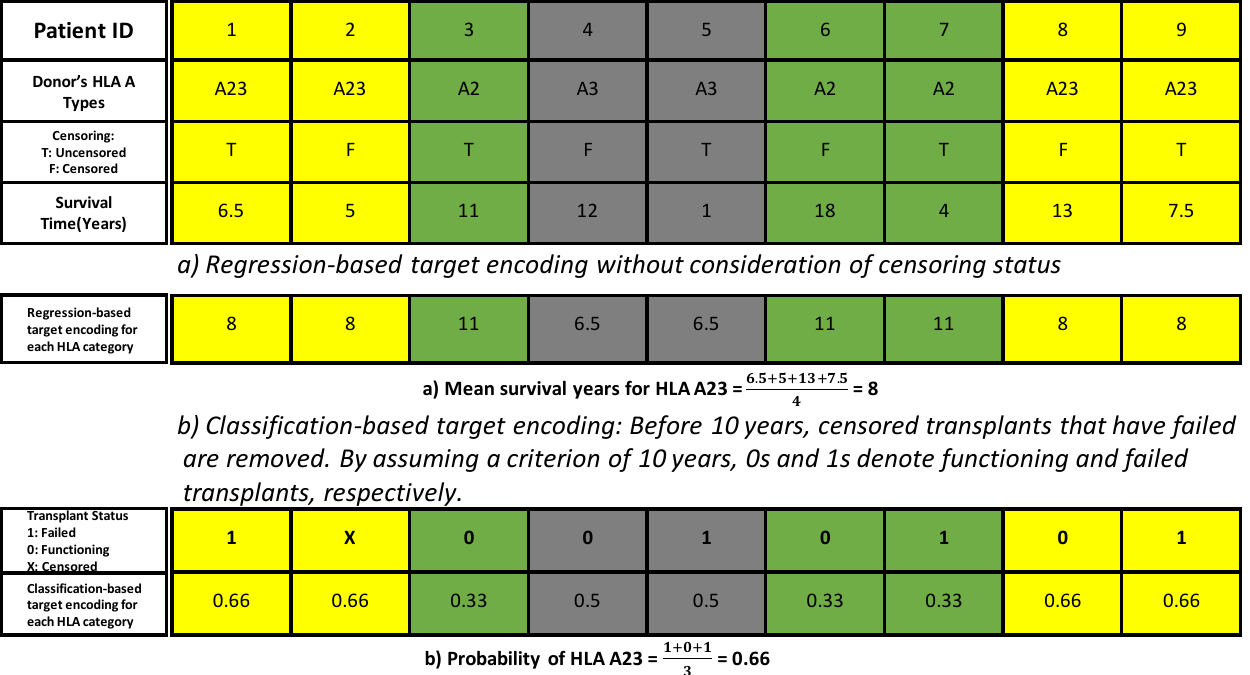}
    \caption{\change{Two examples of target encoding calculation for each HLA type.
    The encodings for the two HLA types for each locus are averaged to form the final target encoding for the locus.}}
    \label{fig: target encoding example}
\end{figure}

To perform the target encoding procedure, we consider the graft survival time in two scenarios: one in which the graft survival time is a continuous target, and the other in which the graft survival time is a binary target. 
Examples of both types of encodings are shown in Figure \ref{fig: target encoding example}.
In the first setting, which we denote as \emph{regression-based target encoding}, the encoding values are simply the average of graft survival times (for uncensored cases) or censoring times (for censored cases) grouped by distinct HLA types. 

In the second setting, which we denote as \emph{classification-based target encoding}, we take into account the censoring status of each transplant. 
By using the number of post transplant years as a criterion, we transform the continuous target to binary values which represent the functioning or failed grafts. 
For determining the encoding values, we eliminate all censored transplants that have not reached the specified number of years. 
For these transplants, we cannot determine whether a graft is still functioning or has failed. 

The encoding for each HLA type is then calculated as follows: 
\begin{equation*}
\text{Encoding at year } t = \frac{\# \text{ of failed transplants at year } t}{\# \text{ of functioning or failed transplants at year } t}.
\end{equation*}
This encoding is simply the average of binarized target (1: failed, 0: functioning) for each HLA type. To investigate the effect of this criterion on our model, we choose one, five, ten, fifteen, and twenty post-transplant years to binarize the target, which result in five distinct HLA type features with distinct encoding values for each type. This also enables us to compare different target encoding approaches. 

To address the ordering problem of each HLA type, we take the average of two encodings calculated for each HLA locus to create a unique encoding for each HLA type regardless of locus.
This allows our encoding to be invariant to the ordering of the two HLA types that a donor or recipient possesses for each locus. 
For example, a donor possessing both A3 and A9 will have their HLA-A target encoding set to the average of the encodings for A3 and A9.

\subsubsection{HLA Pairs}

\change{An HLA pair is the combination of the HLA types of a donor and a recipient. For instance, if a donor possesses HLA-A3 and a recipient possesses A23, the HLA pair (A3, A23) is associated with the transplant. Similar to how we encode HLA types, we can use a one-hot-like encoding for HLA pairs by placing a one in the column for each HLA pair associated with a transplant.}

\begin{figure}[t]
    \centering
    \includegraphics[width = \linewidth]{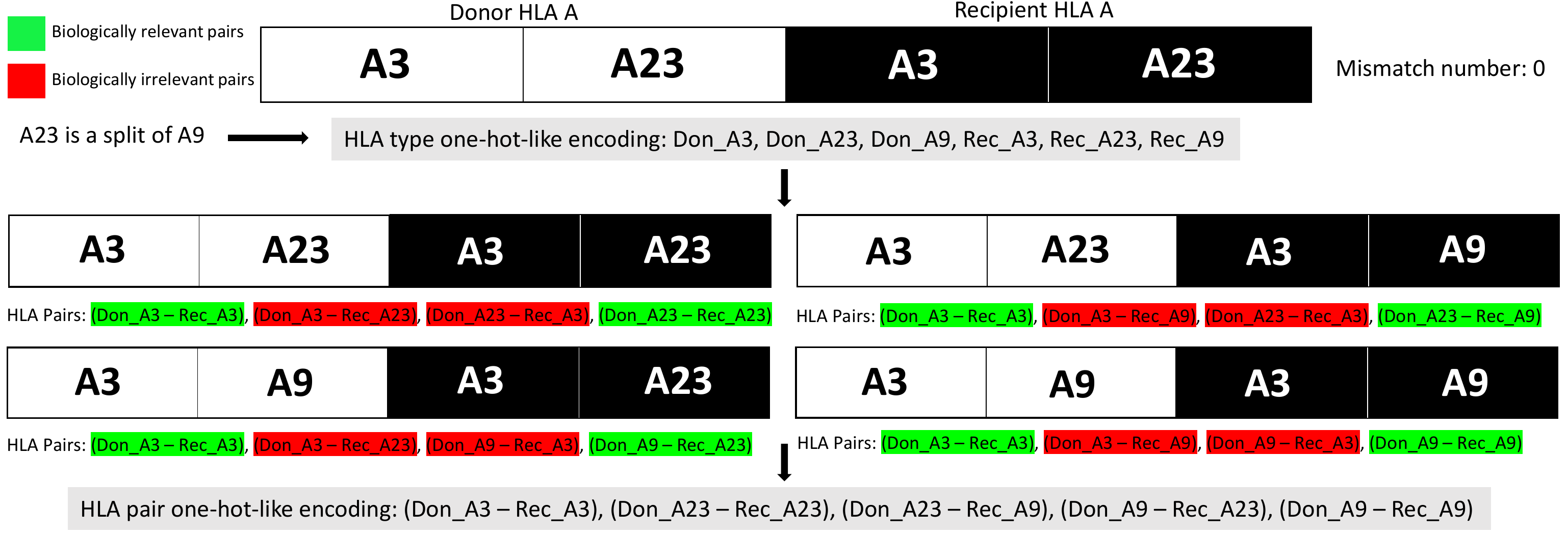}
    \caption{\change{An example of one-hot-like HLA type and pair encoding. 
    The features shown in the gray boxes are encoded as 1's while all other features for the HLA-A locus are encoded as 0's.}}
    \label{fig: HLA type and pair encoding}
\end{figure}

\change{This does not, however, account for the biological mechanisms underlying HLA compatibility. If the donor has HLA types that the recipient does not, the recipient's immune system may reject the transplant. There is no problem if the recipient has HLA types that are not present in the donor, which creates an asymmetry in roles of the donor and recipient HLA types.
As a result, some HLA pairs are not biologically relevant. To distinguish between biologically relevant and irrelevant pairs, it is essential to consider the number of donor-recipient mismatches.
Figure \ref{fig: HLA type and pair encoding} illustrates an instance in which the donor and recipient HLA-As are A3 and A23. Given that A23 is a split of A9, both the donor and the recipient have the HLA types A3, A23, and A9 encoded in our encoding mechanism.

Applying the biological concept of broads and splits into our encoding methodology generates four different cases in pairing the HLAs.  Combining the HLAs of the donor and recipient in each case yields four distinct pairs. In the case where the donor and recipient HLAs are A3 and A23, for instance, there are four possible pairings: (A3, A3), (A3, A23), (A23, A3), and (A23, A3). For this transplant's encoding, we would typically insert one into each column of the four mentioned pairs. However, this does not account for the underlying biological mechanisms of HLA pairs.  

As both of the donor's HLA types are present in the recipient's HLA types, the number of mismatches is zero. Due to this zero mismatch, the mismatched pairs (A3, A23) and (A23, A3) are deemed biologically irrelevant; therefore, the ones in the columns of irrelevant pairs should be replaced with zeros and we also maintain one in the columns of only active HLA pairs (A3, A3), and (A23, A23). Other cases are handled in a similar fashion.}

The HLA pair encoding results in 3,638  features appended to the basic features. 
Due to the large number of features for the HLA pair encoding, we also consider a smaller \emph{frequent pairs} representation where we remove all HLA pairs observed in less than 1,000 transplants, which results in 204 features. 

\subsubsection{All Features}
We consider also a combined feature set by concatenating all of the above feature representations. 
For HLA types, we use the binary one-hot-like encoding. 
The total number of features is 3,894 in the pre-transplant setting, which is dominated by the 3,638 HLA pair features.

\subsection{Survival Analysis Algorithms}
\label{sec:algorithms}

\paragraph{Coxnet}
The Cox Proportional Hazards (Cox PH) model is one of the most widely used models for survival analysis. 
It models the hazard ratio using a weighted linear combination of covariates.
The coefficient vector is estimated by maximizing the partial likelihood. 
We use a Cox PH model with combined $\ell_1$ and $\ell_2$ regularization, known as the elastic net, which leads to the Coxnet model \cite{simon2011regularization}.
The model has 2 hyperparameters: $\lambda$, which controls the strength of regularization, and $r$, which denotes the ratio between the $\ell_1$ and $\ell_2$ penalties. 
We use a grid search with $\lambda$ uniformly distributed on a log scale between $10^{-4}$ and $10^{-2}$ and $r$ uniformly distributed between $0.1$ and $1$.

\paragraph{Random Survival Forest}
Random forest is a bootstrap aggregating (bagging) ensemble learning algorithm with decision trees as base learners.
Ishavan et al.~\cite{ishwaran2008random} proposed the random survival forest (RSF) algorithm that can handle right-censored data.
We use an RSF with $500$ trees and consider random selection of the square root of the number of features for each split. 
We perform a grid search on the maximum depth of each tree in the range $\{5, 10, 15\}$.

\paragraph{Gradient Boosted Regression Trees}
Gradient boosting (GB) is an ensemble learning technique that combines the predictions of many weak learners.
Boosting algorithms 
using survival regression trees as their weak learners have been developed to be used in survival analysis problems \cite{wang2019machine}. 
We use stochastic gradient boosting with $500$ trees using a $50\%$ subsample to fit each tree. 
We perform a grid search on the maximum depth of each tree in the range $\{1, 2, 3\}$.

\section{Empirical Evaluation}
\label{sec:empirical}
To evaluate the accuracy of our predictors, we randomly split the data into 3 sets: $60\%$ training, $20\%$ validation, and $20\%$ testing. 
The validation set is used for hyperparameter tuning. 
For each algorithm, we choose the set of hyperparameters with the highest validation set C-index (see Section \ref{sec:metrics}) and then retrain it on the $80\%$ set containing both the training and validation sets.
We then finally evaluate each algorithm and feature set on the $20\%$ test set, which was initially held out and not used at any point to prevent test set leakage. Furthermore, we repeat the above process of calculating the accuracies of algorithms $10$ times using 10 different splits to avoid drawing conclusions based on a single data split and the potential variance associated with that random split.
Our experiments are conducted using the scikit-survival Python package \cite{polsterl2015fast}.

\subsection{Evaluation Metrics}
\label{sec:metrics}
We consider two metrics to evaluate the accuracy of our survival time predictions.
First, we use Harrell's concordance index (C-index), which is perhaps the most widely used accuracy metric for survival prediction models \cite{harrell1982evaluating}. 
The C-index is merely dependent on the \emph{ordering} of predictions and is calculated by counting all possible pairs of samples and concordant pairs. A pair is a concordant pair if the risk $\eta_i < \eta_j$ and $T_i > T_j$, where $T_i$ is the survival time for patient $i$. 

We also consider the cumulative/dynamic area under receiver operating characteristic curve (AUC) metric that measures how accurately a model can predict the events that happen before and after a specific time $t$ \cite{lambert2016summary}. 
We consider the mean cumulative/dynamic AUC over 5 equally-spaced time points.

\subsection{Statistical Significance Testing}
\label{sec:hypothesis_testing}
\change{Our preliminary results from \cite{nemati21} used a single train, validation, and test split. 
It is difficult to draw conclusions from the results because the observed improvements from incorporating the HLA feature representations are small and may be just due to variance from the single random split.
A major goal of this study is to more definitively evaluate} how our proposed HLA feature representations affect the accuracy of graft survival prediction algorithms. Since the accuracy gain from incorporating the new features may be minor in comparison to our baseline criterion, one might wonder if the gain is due to chance factors like a specific data train and test split. 
To investigate whether this is the case, we conduct our experiments with 10 different data train and test splits, as explained in Section \ref{sec:empirical}. This strategy enables us to conduct an appropriate statistical significance test. 

To compare the test set accuracy of our augmented data to our basic feature set, we use the Wilcoxon signed rank test. The reason for using this test is that we only run each algorithm 10 times, so it is best to avoid making normality assumptions about the population of test set accuracies. In this setting, a non-parametric test like the Wilcoxon signed rank test (rather than a \change{paired} t-test) is a better candidate to make a comparison about the two populations by using pairs of matched samples \cite{conover1999practical}. 
\change{We do also perform a paired t-test to provide a comparison to the results of the Wilcoxon signed rank test.}

To determine whether there is a significant difference between the basic and augmented feature sets, we form the following null and lower-tail alternative hypotheses:
\begin{equation*}
H_{\text{null}}: M_{\text{basic}} = M_{\text{augmented}} \qquad \text{vs.}\qquad H_{\text{alternative}}: M_{\text{basic}} < M_{\text{augmented}},
\end{equation*}
where $M$ denotes the median accuracy metric (either C-index or mean cumulative/dynamic AUC) over the 10 splits. 
We compute the p-values for these tests and compare them to the level of significance $\alpha = 0.05$. 

\change{We consider 6 different augmented feature sets, each of which we compare against the basic features, so we have 6 different hypotheses. 
Since multiple comparison tests are conducted, the Bonferroni correction method is applied to adjust the p-values. The Bonferroni correction is a multiple-comparison correction approach utilized when multiple dependent or independent statistical tests are conducted simultaneously~\cite{abdi2007bonferroni}. To control the excessive occurrence of false positives, which is equivalent to rejecting the null hypothesis, the p-values of each individual comparison must be multiplied by the total number of possible pairwise comparisons between each group to account for the number of comparisons being conducted. 
Therefore, we multiply the p-values of each comparison by 6, the total number of comparisons.}

\section{Results}

\subsection{Effects of Feature Representations}
\label{sec:results_features}

\begin{table}[t]
\setlength{\tabcolsep}{3pt}
\centering
\caption{Average test set C-index accuracy using only pre-transplant covariates across 10 different data splits. The Wilcoxon test p-values are calculated with respect to the basic feature set. \change{Other feature sets also include the basic features.} Best feature set for each predictor is listed in bold.}
\label{tab:c_index_pre}
\begin{tabular}{ccccccc}
\hline
            & \multicolumn{2}{c}{Coxnet}  & \multicolumn{2}{c}{Random Surv.~Forest} & \multicolumn{2}{c}{Gradient Boosting} \\
Feature Set & C-index       & p-value    & C-index          & p-value                & C-index    & p-value \\
\hline
Basic       &0.625          & {\bf --}    &0.634             &  {\bf --}             & 0.636      & {\bf --} \\  
MM (total)  &0.627          & \change{0.006}       &0.635             & \change{0.006}                  & 0.637      & \change{0.012}  \\
MM (A-B-DR) &0.627          & \change{0.006}       &{\bf 0.636}       & \change{0.012}                  & 0.638      & \change{0.006}  \\ 
\change{Types (binary)} &0.626          & \change{0.012}       &0.630             & \change{1}                      & 0.637      & \change{0.006}  \\
Pairs       &{\bf 0.627}    & \change{0.006}       &0.620             & \change{1}                      & 0.637      & \change{0.006}  \\
Freq.~pairs &0.627          & \change{0.006}       &0.631             & \change{1}                      & {\bf 0.638}& \change{0.006}  \\ 
All         &0.627          & \change{0.006}       &0.614             & \change{1}                      & 0.637      & \change{0.006}  \\
\hline
\end{tabular}
\end{table}

\begin{table}[t]
\setlength{\tabcolsep}{3pt}
\centering
\caption{Average test set mean cumulative/dynamic AUC accuracy using only pre-transplant covariates across 10 different data splits. The Wilcoxon test p-values are calculated with respect to the basic feature set. \change{Other feature sets also include the basic features.} Best feature set for each predictor is listed in bold.}
\label{tab:mean_auc_pre}
\begin{tabular}{ccccccc}
\hline
            & \multicolumn{2}{c}{Coxnet}  & \multicolumn{2}{c}{Random Surv.~Forest} & \multicolumn{2}{c}{Gradient Boosting} \\
Feature Set                 & Mean AUC      & p-value         & Mean AUC             & p-value      & Mean AUC        & p-value \\
\hline
Basic                       &0.635          & {\bf --}        &0.655                  &{\bf --}     & 0.653           &{\bf --}  \\  
MM (total)                  &{\bf 0.641}    &\change{0.006}            &{\bf 0.659}            &\change{0.006}        & {\bf 0.658}     &\change{0.006}   \\
MM (A-B-DR)                 &0.641          &\change{0.006}            &0.659                  &\change{0.018}        & 0.658           &\change{0.006}   \\ 
\change{Types (binary)}     &0.635          &\change{1}                &0.654                  &\change{1}            & 0.654           &\change{1}   \\
Pairs                       &0.640          &\change{0.006}            &0.651                  &\change{1}            & 0.655           &\change{0.012}   \\
Freq.~pairs                 &0.640          &\change{0.006}            &0.658                  &\change{0.006}        & 0.657           &\change{0.006}   \\ 
All                         & 0.640         &\change{0.006}            &0.645                  &\change{1}            & 0.657           &\change{0.006}   \\
\hline
\end{tabular}
\end{table}

The two main research questions are both centered around the effects of incorporating HLA information into graft survival time prediction. 
From the results in Tables \ref{tab:c_index_pre} and \ref{tab:mean_auc_pre}, notice that incorporating HLA information almost always results in an improvement in prediction accuracy in the pre-transplant prediction setting. 
The amount of improvement compared to the basic features varies for the differing feature representations and evaluation metrics. 
\change{In addition to the mean accuracy metrics, the tables also show the p-values from the Wilcoxon signed rank test. 
The differences between the C-indices for each feature set and the C-index for the basic features on each of the 10 data splits along with p-values from a paired t-test are shown in \ref{sec:appendix_features}.}

\paragraph{HLA Mismatches}
In all cases, adding HLA MM (either total or separate A-B-DR MM) improves the accuracy of the predictive models.
We notice minimal differences in accuracy from including total MM and A-B-DR MM. 
Additionally, the p-values suggest that the improvement is statistically significant at the $\alpha = 0.05$ level. 

\change{The maximum improvement in prediction accuracy observed across both evaluation metrics and all three algorithms is obtained from the MM (total) feature set. The relative improvement of mean AUC from 0.635 for the basic features to 0.641 when including MM (total) is about 1\%.}

\paragraph{HLA Types}
\change{We first consider the binary encoding for HLA types. Results for the target encoding are shown in Section \ref{subsub: target_encoding_discussion}.}

For Coxnet and gradient boosting, including HLA types resulted in better accuracy than the basic features. 
However, for all of the predictors, including HLA types resulted in worse accuracy than including HLA MM. 
Since the Coxnet is linear in the features, it cannot learn interactions between features, and thus, cannot learn compatibilities between different HLA types, so this result is not too surprising for Coxnet. 
On the other hand, the tree-based predictors are non-linear and should be able to learn donor-recipient HLA compatibilities, so it is somewhat surprising that RSF and GB also perform worse. 
We discuss some possibilities below when \change{considering HLA pairs. 

For RSF, notice that including HLA types leads to a C-index even lower than just using the basic feature set. 
This does not happen with Coxnet or GB and leads us to consider target encoding approaches for the HLA types when using RSF, which we discuss in Section \ref{subsub: target_encoding_discussion}.}

The p-values for Coxnet and GB suggest that, when HLA types are added to basic features, there is a statistically significant improvement in test set accuracy when measured using the C-index, but not the mean AUC. 
This may be due to the algorithms' hyperparameters being optimized using C-index rather than mean AUC.

\paragraph{HLA Pairs}
Unlike with HLA MM, the results with HLA pairs vary by model. 
The inclusion of all HLA pairs benefits the Coxnet more than any other feature.
Since it is linear in the features, it requires HLA pair  features in order to learn compatibilities between donor and recipient HLAs. 
It is also robust to overfitting in high dimensions due to the elastic net penalty.
Thus, it is not surprising the HLA pairs lead to the highest C-index for Coxnet.

On the other hand, the nonlinear predictors behave differently as they see a minimal gain (GB) or even a significant decrease (RSF) in accuracy from the inclusion of all HLA pairs. 
This indicates that the high dimensionality may cause a problem for tree-based predictors, particularly for the RSF.
The high dimensionality results from the one-hot-like encoding mechanism we are using for HLA pairs, which can be disadvantageous for trees because it splits a single categorical variable into multiple variables, potentially requiring many splits for a single categorical variable with a large number of categories. 
When restricting to just the most frequent HLA pairs, resulting in a much smaller number of HLA pair features (180 compared to over 3,600), the accuracy of GB now increases rather than decreases, and results are mixed for RSF (decrease in C-index but increase in mean AUC). 

\paragraph{All}
The accuracy when all features are included seems to be similar to that of including all HLA pairs, which contribute the highest number of features. 
Both Coxnet and GB have statistically significant improvements when using all features compared to the basic feature set.

\subsection{Effects of Target Encoding}
\label{subsub: target_encoding_discussion}

\begin{table}[t]
\setlength{\tabcolsep}{3pt}
\centering
\caption{RSF type feature set mean survival prediction accuracy using only pre-transplant covariates across 10 different data splits for different \change{HLA type encodings}. Best encoding version is listed in bold for each metric.}
\label{tab:c_ind._meanAUC_pval.}
\begin{tabular}{ccccccc}
\hline
\change{HLA Type} Encoding      & C-index        & p-value & Mean AUC    & p-value  \\
\hline     
\change{Binary}               & \change{0.630}          & \change{--}      &     \change{0.654}       & \change{--} \\
\change{Target (Regression-based)}           & 0.633          & \change{0.006}   & 0.652       & \change{1}                   \\
\change{Target (Classification: 1 year)}  & 0.630          & \change{1}         & 0.651       & \change{1}              \\
\change{Target (Classification: 5 years)}  & 0.633          & \change{0.006}   & 0.652       & \change{1}                                          \\
\change{Target (Classification: 10 years)} & 0.633          & \change{0.006}   & 0.653       & \change{1}                                          \\
\change{Target (Classification: 15 years)} & 0.633          & \change{0.006}   & 0.653       & \change{1}                                          \\
\change{Target (Classification: 20 years)} & {\bf0.634}     & \change{0.006}   & {\bf0.655}  & \change{0.316}                                          \\
\hline
\end{tabular}
\end{table}

We saw in the previous section that the addition of binary-encoded HLA types has a negative impact on RSF's performance due to its high dimensionality. 
\change{We compare the accuracy of RSF using different target encodings for the HLA types to that of the binary encoding. To formalize the comparison as a statistical test, we consider the following null and alternative hypotheses:
\begin{equation*}
H_{\text{null}}: M_{\text{binary}} = M_{\text{target}} \qquad \text{vs.}\qquad H_{\text{alternative}}: M_{\text{binary}} < M_{\text{target}},
\end{equation*}
where $M$ denotes the median accuracy metric (C-index or mean AUC) over the 10 data splits. 
The p-values are computed using the Wilcoxon signed rank test with Bonferroni correction in the same manner as described in Section \ref{sec:hypothesis_testing}.
The differences between the C-indices of the target encodings and the binary encoding are shown in \ref{sec:appendix_target}.}

The type feature set test accuracy using the proposed regression and classification-based target encoding approaches is shown in Table \ref{tab:c_ind._meanAUC_pval.}. The C-index values for the binary and target-encoded type set test accuracy suggests that both target encoding approaches can improve RSF's predictive power.  The p-values also indicate that there is a statistically significant improvement for the C-index of target encoding (except for the 1-year classification). 
The mean AUC, on the other hand, does not improve, which could again be due to the hyperparameters being optimized using C-index.

\subsection{Prediction with Post-transplant Covariates}
When we include also the post-transplant covariates, for all models, the C-index and mean cumulative/dynamic AUC improve by about 0.03-0.04 compared to using only pre-transplant covariates. 
The highest C-indices are 0.667, 0.676, and 0.678 for Coxnet, random survival forest and gradient boosting, respectively, as shown in Table \ref{tab:c_index_pre_post}.
The results indicate that integrating post-transplant covariates tremendously helps the survival prediction algorithms improve their accuracy, as one might expect.

The trends across HLA feature representations are roughly the same as in the pre-transplant case, although the relative improvement in accuracy when including the HLA features is slightly lower compared to using only pre-transplant features. 
This is reasonable because the post-transplant covariates carry information about how the recipient's immune system is responding to the transplant, which is what the HLA features aim to predict in the pre-transplant setting.

\begin{table}[t]
\setlength{\tabcolsep}{3pt}
\centering
\caption{Survival prediction accuracy using pre- and post-transplant covariates. Best feature set for each predictor and each metric is listed in bold.}
\label{tab:c_index_pre_post}
\begin{tabular}{ccccccc}
\hline
            & \multicolumn{2}{c}{Coxnet} & \multicolumn{2}{c}{Random Surv.~Forest} & \multicolumn{2}{c}{Gradient Boosting} \\
Feature Set & C-index     & Mean AUC     & C-index     & Mean AUC     & C-index     & Mean AUC \\
\hline
Basic       & 0.665       & 0.673        & 0.676        & 0.690       & 0.677      & 0.687  \\
MM (total)  & 0.667       &{\bf 0.678}   & 0.676        & 0.692       & 0.678      &{\bf 0.690}   \\
MM (A-B-DR) & 0.667       & 0.678        &{\bf 0.676}   &{\bf 0.693}  &{\bf 0.678} & 0.689       \\ 
Types       & 0.666       & 0.672        & 0.667        & 0.687       & 0.678      & 0.686      \\
Pairs       & 0.667       & 0.677        & 0.658        & 0.681       & 0.677      & 0.688      \\
Freq.~pairs & 0.667       & 0.677        & 0.669        & 0.687       & 0.678      & 0.689      \\ 
All         &{\bf 0.667}  &  0.677       & 0.654        & 0.680       & 0.677      & 0.687       \\
\hline
\end{tabular}
\end{table}

\section{Related Work}
A broad group of studies has used data-driven statistical models to predict graft survival times or measure risk factors' impact on graft survival. 
Prior work includes 
multivariate analysis using Cox proportional hazards (Cox PH) models with a small number of covariates~\cite{rao2009comprehensive,wolfe2009predictability,ashby2017kidney}. There has been more recent work on machine learning-based survival analysis applied to kidney transplantation, including an ensemble model that combines Cox PH models with random survival forests \cite{mark2019using} and a deep learning-based approach \cite{luck2017deep}.

Our results compare favorably to prior studies \cite{rao2009comprehensive,wolfe2009predictability,ashby2017kidney,luck2017deep} using the same SRTR data we use in this study. 
Each study differs in inclusion criteria, time duration, and several other factors that prevent a direct comparison; however, we include their reported results here for reference. 
Two older studies \cite{rao2009comprehensive} and \cite{wolfe2009predictability} using Cox PH models without regularization achieved C-indices of $0.62$ and $0.61$, respectively. 
A more recent study also using a Cox PH model with only pre-transplant covariates \cite{ashby2017kidney} including HLA MM achieved a C-index of $0.64$; however, their study included both living and deceased donors while ours considers only deceased donors. 
Transplant outcomes with living donors are much more favorable 
\cite{ashby2017kidney}, which may result in easier prediction. 
Another recent study \cite{luck2017deep} used a deep learning approach applied to both pre- and post-transplant covariates to achieve a C-index of $0.655$, less than the $0.676$ we achieved. 

Several other recent studies have focused on prediction of patient survival rather than graft survival, with \cite{li2016predicting} and \cite{mark2019using} achieving C-indices of $0.70$ and $0.724$, respectively. 
Prediction of patient survival is much easier than prediction of graft survival, which we focus on in this paper. 
For example, \cite{wolfe2009predictability} considered both patient and graft survival and achieved a C-index of $0.68$ for patient survival compared to $0.61$ for graft survival. 
We also argue that graft survival is the more relevant clinical endpoint, as a patient who survives a transplant but suffers a graft failure will require a re-transplant and returns to the waiting list.

\section{Significance and Impact}
Transplantation outcome prediction is instrumental for clinical decision-making, as well as allocation policy development. The kidney allocation policy by the OPTN was developed to encourage fairness (equal access to treatment) and effectiveness (the longest predicted survival) \cite{israni2014new} in transplantation. Informed clinical decision making allows for avoidance of high-risk transplants and thus reduces number of graft losses. However, accurate prediction of transplant outcomes remains a daunting challenge due to the high complexity of human biology. 

\change{In addition, failure to account for complexities of HLA results in unintended consequences in transplantation. As such, OPTN's good intention to promote HLA matching initially resulted in de facto discrimination against African Americans, whose HLA gene locus is highly diverse and who therefore were not selected for transplantation as frequently as Caucasians and other races \cite{young2000renal}. The requirement for HLA matching was later relaxed, but the problem of racial disparities in access to high-quality transplants persists until today \cite{taber2017outcome}. By modifying our approach to HLA immunogenicity quantification, adding biologically-relevant representations of HLA, we attempt to build improved models for transplant outcome prediction, which may help address the pressing problem of poor long-term transplantation outcomes. 

Addition of HLA features improves our predictive model and presents clinical interest for two reasons. First, physicians are most comfortable making decisions with HLA information at hand. There is a growing consensus in the transplantation field that HLA is a critical consideration for pre-transplant patient evaluation \cite{poggio2021long}. In the U.S., nationwide sharing of fully HLA-matched kidneys is mandated in certain situations, and transplant centers typically require labs to provide HLA information before a crossmatch (a final pre-transplant test). Therefore, clinicians, governmental entities, and payers who are interested in predicting transplantation outcome are typically interested in making sure HLA compatibility is factored into the model.

Second, due to the large size of the transplant waiting list and exorbitant cost of pre-transplant kidney replacement therapy, even a small improvement in post-transplant outcomes would result in large economic and social impact over time, as was described in simulations by Segev et al.~\cite{segev2005kidney}. They showed that as much as \$750 million could be saved if transplant rates were to improve by 5.7\% in a 4,000 patient pool. It would require a separate study to quantify the impact of a 1\% increase in prediction accuracy on long-term graft survival, however, it is reasonable to think that implementation of improved predictive models in transplant allocation would result in improvement in transplant survival, with downstream societal impact. Our findings are thus useful for assisting clinical decision making aimed at improving long-term allograft survival.}

\section*{Acknowledgements}
\change{We thank Rong Liu for his assistance with the statistical significance tests and the anonymous reviewers for their suggestions to improve the paper.}

The research reported in this publication was supported by the National Library of Medicine of the National Institutes of Health under Award Number R01LM013311 as part of the NSF/NLM Generalizable Data Science Methods for Biomedical Research Program. The content is solely the responsibility of the authors and does not necessarily represent the official views of the National Institutes of Health.

The data reported here have been supplied by the Hennepin Healthcare Research Institute (HHRI) as the contractor for the Scientific Registry of Transplant Recipients (SRTR). The interpretation and reporting of these data are the responsibility of the author(s) and in no way should be seen as an official policy of or interpretation by the SRTR or the U.S.~Government.
Notably, the principles of the Helsinki Declaration were followed.

\appendix

\change{
\section{Evaluation Metrics Across Data Splits}
\label{sec:appendix_splits}

\subsection{Effects of Feature Representations}
\label{sec:appendix_features}

\begin{figure}[tp]
    \centering
    \includegraphics[scale = 0.65]{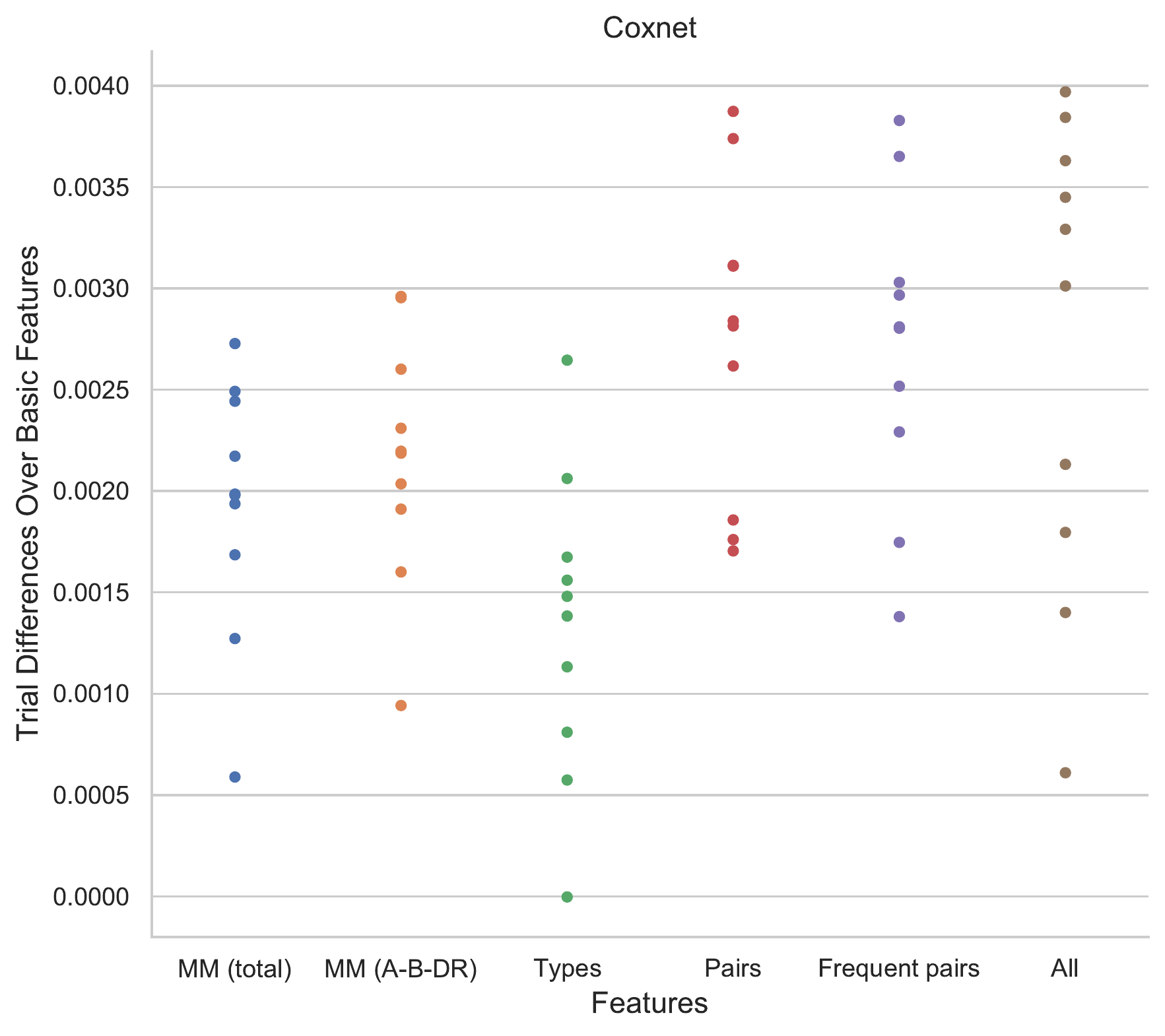}
    \caption{ \change{Difference between C-index of each feature set and C-index of basic feature set on all 10 data splits for the Coxnet. For the Types feature set, 9 of 10 splits result in improved C-index. For all other feature sets, all 10 splits result in improved C-index.}}
    \label{fig:trial differences coxnet one hot}
\end{figure}

\begin{figure}[tp]
    \centering
    \includegraphics[scale = 0.65]{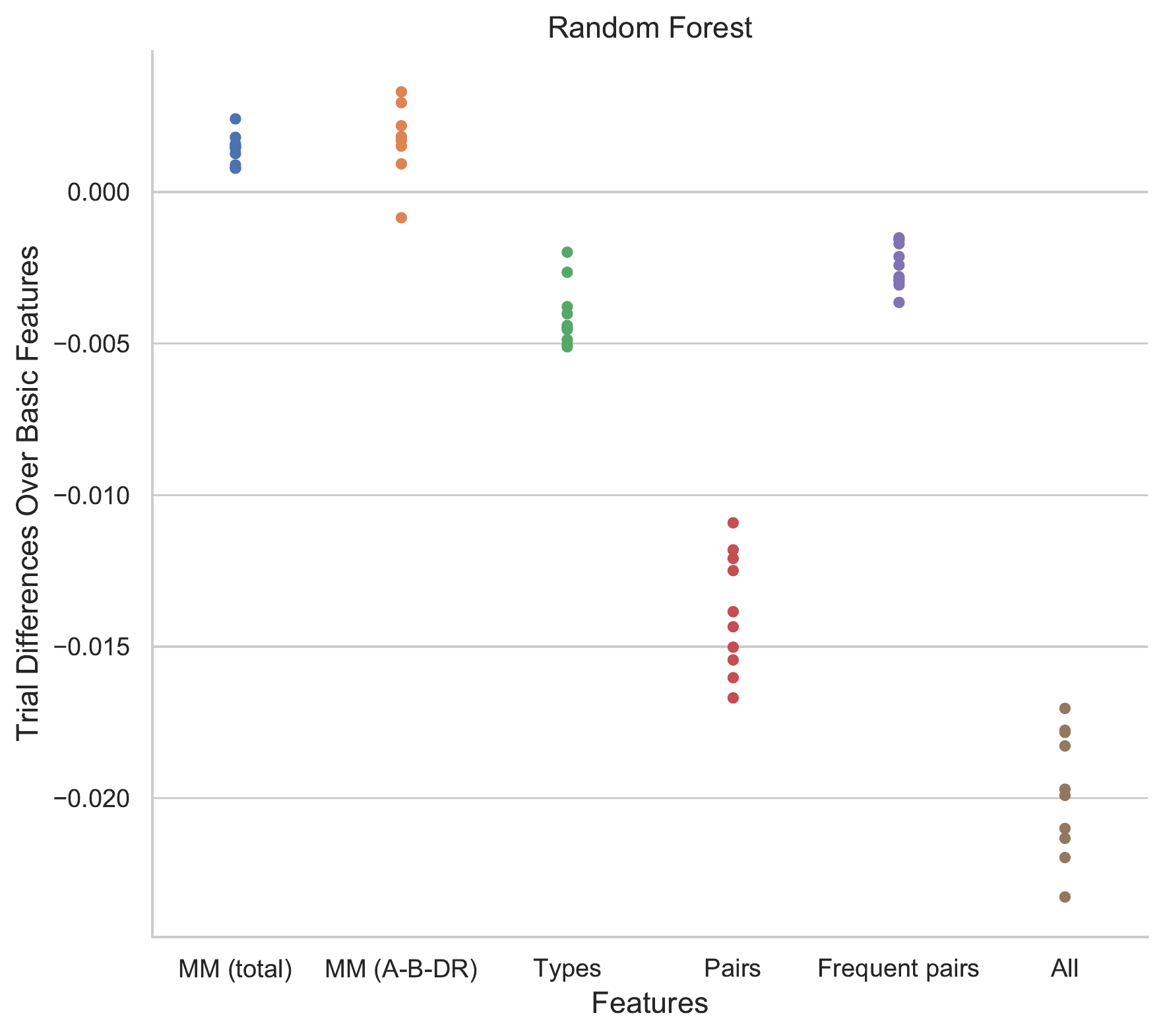}
    \caption{ \change{Difference between C-index of each feature set and C-index of basic feature set on all 10 data splits for Random Survival Forest. MM (total) shows improved C-index in all 10 splits, while MM (A-B-DR) shows improved C-index in 9 of 10 splits. All other feature sets result in decreased C-index in all 10 splits.}}
    \label{fig:trial differences rf one hot}
\end{figure}

\begin{figure}[tp]
    \centering
    \includegraphics[scale = 0.65]{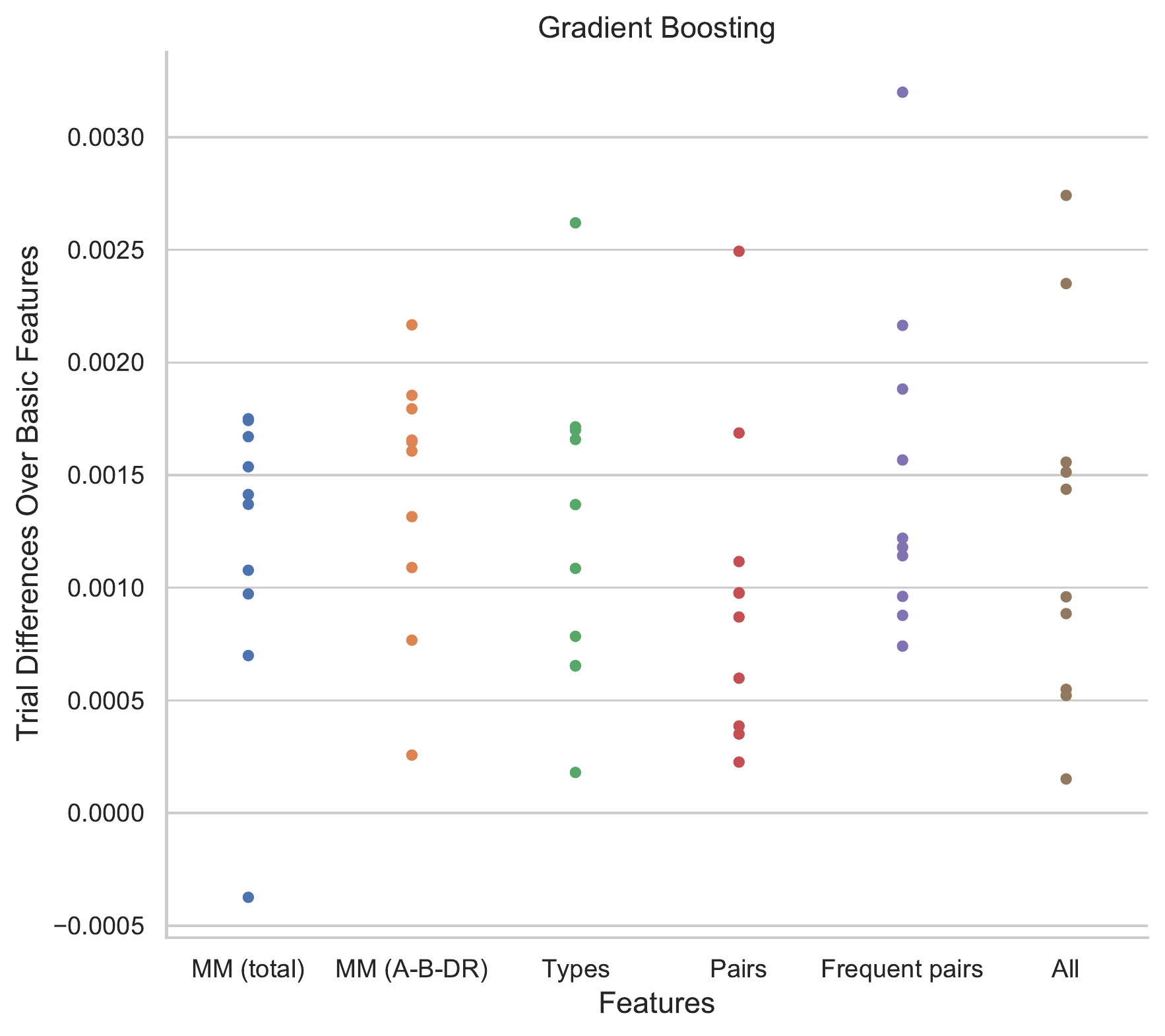}
    \caption{ \change{Difference between C-index of each feature set and C-index of basic feature set on all 10 data splits for Gradient Boosting. MM (total) shows improved C-index in 9 of 10 splits. All other feature sets show improved C-index in all 10 splits.} }
    \label{fig:trial differences gb one hot}
\end{figure}

The difference between the C-index of each feature set ($M_{\text{augmented}}$) and the C-index of the basic feature set ($M_{\text{basic}}$) is shown in Figures \ref{fig:trial differences coxnet one hot}-\ref{fig:trial differences gb one hot} for Coxnet, random survival forest, and gradient boosting, respectively. 
Notice that, for all the cases with p-value 0.006 from the Wilcoxon signed rank test, $M_{\text{augmented}} > M_{\text{basic}}$ for all 10 data splits. 
While the improvement in C-index from adding HLA features is small, it is consistent across the 10 data splits, leading to the low p-values in Table \ref{tab:c_index_pre}.

\begin{table}[tp]
\setlength{\tabcolsep}{3pt}
\centering
\caption{\change{Comparison of Wilcoxon signed rank test and paired t-test p-values for C-index}}
\label{tab:wilcoxon t-test c-index comparison}
\begin{tabular}{ccccccc}
\hline
            & \multicolumn{2}{c}{\change{Coxnet}} & \multicolumn{2}{c}{\change{Random Surv.~Forest}} & \multicolumn{2}{c}{\change{Gradient Boosting}} \\
            & \change{Wilcoxon}         & \change{t-test}       & \change{Wilcoxon}     & \change{t-test}    & \change{Wilcoxon}        & \change{t-test} \\
\hline
\change{MM (total)}  &\change{0.006}             &\change{0}             &\change{0.006}        &\change{0}           &\change{0.012}            &\change{0.001}         \\
\change{MM (A-B-DR)} &\change{0.006}             &\change{0}             &\change{0.012}        &\change{0.003}       &\change{0.006}            &\change{0}        \\ 
\change{Types}       &\change{0.012}             &\change{0.001}         &\change{1}            &\change{1}           &\change{0.006}            &\change{0.001}       \\
\change{Pairs}       &\change{0.006}             &\change{0}             &\change{1}            &\change{1}           &\change{0.006}            &\change{0.005}       \\
\change{Freq.~pairs} &\change{0.006}             &\change{0}             &\change{1}            &\change{1}           &\change{0.006}            &\change{0}      \\ 
\change{All}         &\change{0.006}             &\change{0}             &\change{1}            &\change{1}           &\change{0.006}            &\change{0.003}        \\
\hline
\end{tabular}
\end{table}

\begin{table}[tp]
\setlength{\tabcolsep}{3pt}
\centering
\caption{\change{Comparison of Wilcoxon signed rank test and paired t-test p-values for mean AUC}}
\label{tab:wilcoxon t-test mean AUC comparison}
\begin{tabular}{ccccccc}
\hline
            & \multicolumn{2}{c}{\change{Coxnet}} & \multicolumn{2}{c}{\change{Random Surv.~Forest}} & \multicolumn{2}{c}{\change{Gradient Boosting}} \\
            & \change{Wilcoxon}         & \change{t-test}        &\change{Wilcoxon}          & \change{t-test}     & \change{Wilcoxon}        & \change{t-test} \\
\hline
\change{MM (total)}  &\change{0.006}             &\change{0}              &\change{0.006}             &\change{0}           &\change{0.006}            &\change{0}         \\
\change{MM (A-B-DR)} &\change{0.006}             &\change{0}              &\change{0.018}             &\change{0.007}       &\change{0.006}            &\change{0}        \\ 
\change{Types}       &\change{1}                 &\change{1}              &\change{1}                 &\change{1}           &\change{1}                &\change{1}       \\
\change{Pairs}       &\change{0.006}             &\change{0}              &\change{1}                 &\change{1}           &\change{0.012}            &\change{0.001}       \\
\change{Freq.~pairs} &\change{0.006}             &\change{0}              &\change{0.006}             &\change{0.001}       &\change{0.006}            &\change{0}       \\ 
\change{All}         &\change{0.006}             &\change{0}              &\change{1}                 &\change{1}           &\change{0.006}            &\change{0}        \\
\hline
\end{tabular}
\end{table}

In Tables \ref{tab:wilcoxon t-test c-index comparison} and \ref{tab:wilcoxon t-test mean AUC comparison}, we compare the p-values computed by the Wilcoxon signed rank test (the ones shown in Tables \ref{tab:c_index_pre} and \ref{tab:mean_auc_pre}, respectively) with those computed using a paired t-test. 
The p-values from the paired t-test are lower, suggesting that the Wilcoxon signed rank test is more conservative about rejecting null hypotheses.

\newpage
\subsection{Effects of Target Encoding}
\label{sec:appendix_target}

\begin{figure}[tp]
    \centering
    \includegraphics[scale = 0.65]{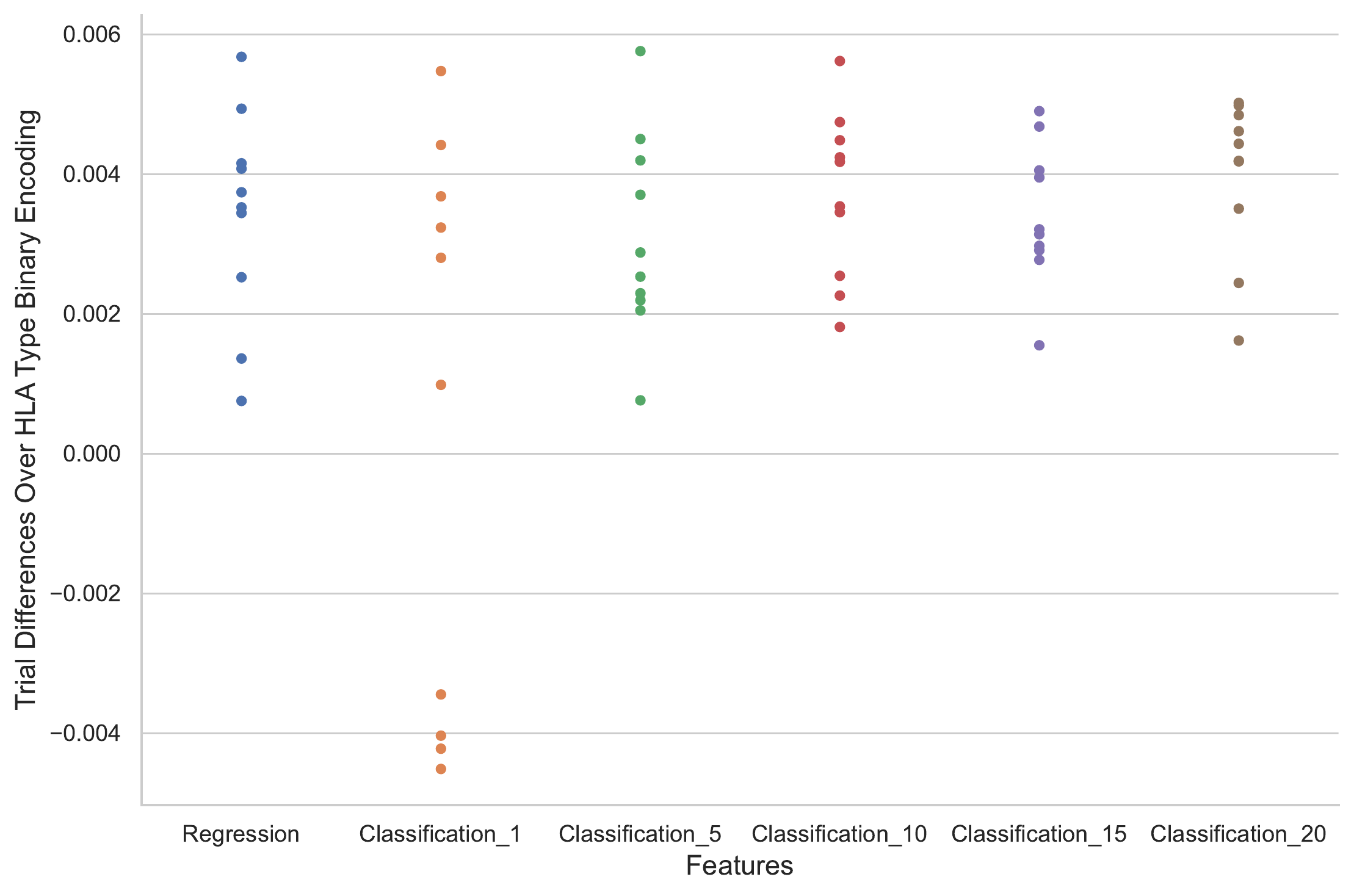}
    \caption{ \change{Difference between C-index of each target encoding approach and C-index of binary encoding of Types feature set on all 10 data splits for random survival forest.
    All encodings except the classification-based encoding at 1 year (Classification\_1) show improved C-index in all 10 splits.}  }
    \label{fig: encoding trial differences rf}
\end{figure}

The difference between the C-index of each target encoding for HLA types ($M_{\text{target}}$) and the C-index of the binary encoding ($M_{\text{binary}}$) is shown in Figure \ref{fig: encoding trial differences rf} for random survival forest. 
The improvement offered by target encoding is consistent across all approaches except for the classification-based encoding at 1 year.
}

 \bibliographystyle{abbrv} 
 \bibliography{References2.bib}

\end{document}